\documentclass[10pt,twocolumn,letterpaper]{article}

\usepackage{cvpr}
\usepackage{times}
\usepackage{epsfig}
\usepackage{graphicx}
\usepackage{amsmath, bm}
\usepackage{amssymb}
\usepackage{epstopdf}
\usepackage{multirow}

\usepackage[breaklinks=true,colorlinks,bookmarks=false]{hyperref}

\cvprfinalcopy 

\def\cvprPaperID{2180} 

\def\Fig#1{{Fig.\ \ref{fig:#1}}}
\def\Eq#1{{Eq.\ \ref{eq:#1}}}
\def\Table#1{{Table \ref{tbl:#1}}}

\ifcvprfinal\fi
\begin{document}

\title{Holistic and Comprehensive Annotation of Clinically Significant Findings on Diverse CT Images: Learning from Radiology Reports and Label Ontology}

\author{Ke Yan$^1$, Yifan Peng$^2$, Veit Sandfort$^1$, Mohammadhadi Bagheri$^1$, Zhiyong Lu$^2$, Ronald M.\ Summers$^1$\\
	$^1$ Imaging Biomarkers and Computer-Aided Diagnosis Laboratory, Clinical Center\\
	$^2$ National Center for Biotechnology Information, National Library of Medicine\\
	$^{1,2}$ National Institutes of Health, Bethesda, MD 20892\\
{\tt\small \{ke.yan, yifan.peng, veit.sandfort, mohammad.bagheri, zhiyong.lu, rms\}@nih.gov}
}

\maketitle

\begin{abstract}

In radiologists' routine work, one major task is to read a medical image, e.g., a CT scan, find significant lesions, and describe them in the radiology report. In this paper, we study the lesion description or annotation problem. Given a lesion image, our aim is to predict a comprehensive set of relevant labels, such as the lesion's body part, type, and attributes, which may assist downstream fine-grained diagnosis. 
To address this task, we first design a deep learning module to extract relevant semantic labels from the radiology reports associated with the lesion images. With the images and text-mined labels, we propose a lesion annotation network (LesaNet) based on a multilabel convolutional neural network (CNN) to learn all labels holistically. Hierarchical relations and mutually exclusive relations between the labels are leveraged to improve the label prediction accuracy. The relations are utilized in a label expansion strategy and a relational hard example mining algorithm. We also attach a simple score propagation layer on LesaNet to enhance recall and explore implicit relation between labels. Multilabel metric learning is combined with classification to enable interpretable prediction. 
We evaluated LesaNet on the public DeepLesion dataset, which contains over 32K diverse lesion images.
Experiments show that LesaNet can precisely annotate the lesions using an ontology of 171 fine-grained labels with an average AUC of 0.9344. The labels of DeepLesion and the code have been released \footnote{\url{https://github.com/rsummers11/CADLab/tree/master/LesaNet}}.

\end{abstract}

\section{Introduction}
\label{sec:intro}

In recent years, there has been remarkable progress on computer-aided diagnosis (CAD) based on medical images, especially with the help of deep learning technologies \cite{Litjens2017survey, Shin2016tmi}. Lesion classification is one of the most important topics in CAD. Typical applications include using medical images to classify the type of liver lesions and lung tissues \cite{Diamant2016liver, Hofmanninger2015cvpr, Shin2016tmi}, to describe the fine-grained attributes of pulmonary nodules and breast masses \cite{Chen2017lung, Ravishankar2016breast}, and to predict their malignancy \cite{Cheng2016Breast, Ferreira2017nodule}. However, existing studies on this topic usually focus on certain body parts (lung, breast, liver, etc.)\ and attempt to distinguish between a limited set of labels. Hence, many clinically meaningful lesion labels covering different body parts were not explored yet. Besides, in practice, multiple labels can be assigned per lesion and are often correlated.

\begin{figure}[]
	\begin{center}
		\includegraphics[width=\linewidth,trim=150 20 175 30, clip]{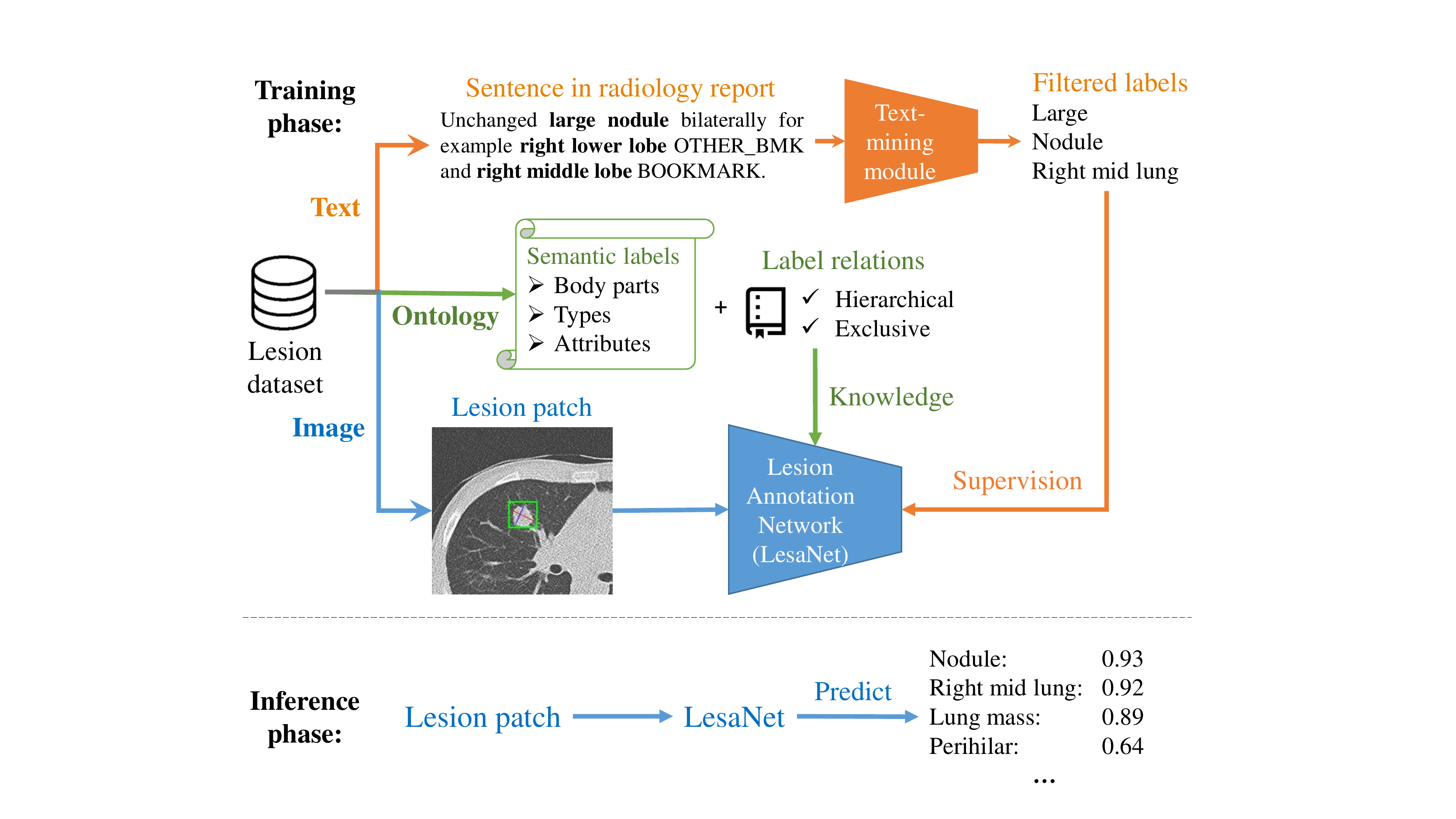} 
	\end{center}
	\caption{The overall framework. We propose lesion annotation network (LesaNet) to predict fine-grained labels to describe diverse lesions on CT images. The training labels are text-mined from radiology reports. Label relations are utilized in learning.}
	\label{fig:overall_framework}
\end{figure}
In this paper, we tackle a more general and clinically useful problem to mimic radiologists. When an experienced radiologist reads a medical image such as a computed tomography (CT) scan, he or she can detect all kinds of lesions in various body parts, identify the lesions' detailed information including its associated body part, type, and attributes, and finally link these labels to the predefined ontology. We aim to develop a new framework to predict these semantic labels holistically (jointly learning all labels), so as to go one step closer to the goal of ``learning to read CT images''. In brief, we wish the computer to recognize where, what, and how the lesion is, helping the user comprehensively understand it. We call this task lesion annotation due to its analogy to the multilabel image annotation/tagging problem in general computer vision literature \cite{Zhang2014Survey}.

To learn to annotate lesions, a large-scale and diverse dataset of lesion images is needed. Existing lesion datasets \cite{Clark2013TCIA, Setio2017LUNA} are typically either too small or less diverse. Fortunately, the recently-released DeepLesion dataset \cite{Yan2018DeepLesion, Yan2018LesGraph} has largely mitigated this limitation. It contains bounding-boxes of over 32K lesions from a variety of body parts on CT images. However, there are no fine-grained semantic labels given for each lesion in DeepLesion. Manual annotation is tedious, expensive, and not scalable, not to mention it requires experts with considerable domain knowledge. Inspired by recent studies \cite{Wang2017ChestXray8, Tang2018attention}, we take an automatic data mining approach to extract labels from radiology reports. Reports contain rich but complex information about multiple findings in the medical image. In the course of interpreting a CT scan, a radiologist may manually annotate a lesion in the image, and place a hyperlink to the annotation (a ``bookmark'') in the report. We first locate the sentence with bookmark in the report that refers to a lesion, then extract labels from the sentence. We defined a fine-grained ontology based on the RadLex lexicon \cite{Langlotz2006RadLex}. This process is entirely data-driven and requires minimal manual effort, thus can be easily employed to build large datasets with rich vocabularies. Sample lesion image, sentence, and labels can be found in \Fig{overall_framework}.

We propose a LESion Annotation NETwork (LesaNet) to predict semantic labels given a lesion image of interest. This lesion annotation task is treated as a multilabel image classification problem \cite{Zhang2014Survey}. Despite extensive previous studies \cite{Dong2018Imbalanced, Gong2013WARP, Hu2016Relation, Wang2016CNNRNN, Li2017Pairwise}, our problem is particularly challenging due to several reasons: {\bf 1)} Radiology reports are often in the format of free-text, so extracted labels can be noisy and incomplete \cite{Wang2017ChestXray8}. {\bf 2)} Some labels are difficult to distinguish or learn, e.g.\ adjacent body parts, similar types, and subtle attributes. {\bf 3)} The labels are highly imbalanced and long-tailed. To tackle these challenges, we present the framework shown in \Fig{overall_framework}. First, we reduce the noise in training labels by a text-mining module. The module analyzes the report to find the labels relevant to the lesion-of-interest. Second, we build an ontology which includes the hierarchical hyponymy and mutually exclusive relations between the labels. With the hierarchical relations, we apply a label expansion strategy to infer the missing parent labels. The exclusive relations are used in a relational hard example mining (RHEM) algorithm to help LesaNet learn hard cases and improve precision. Third, we also attach a simple score propagation layer to enhance recall, especially for rare labels. Finally, metric learning is incorporated in LesaNet to not only improve classification accuracy but also enable prediction interpretability.

The main contributions of this work includes the following: {\bf 1)} We study the holistic lesion annotation problem and propose an automatic learning framework with minimum manual annotation effort;  {\bf 2)} An algorithm is proposed to text-mine relevant labels from radiology reports; {\bf 3)} We present LesaNet, an effective lesion annotation algorithm that can also be adopted in other multilabel image classification problems; and {\bf 4)} To leverage the ontology-based medical knowledge, we incorporate label relations in LesaNet.

\section{Related Work}
\label{sec:relWork}

{\bf Medical image analysis with reports:} Annotating medical images is tedious and requires considerable medical knowledge. To reduce manual annotation burden, some researchers leveraged the rich information contained in associated radiology reports. Disease-related labels have been mined from reports for classification and weakly-supervised localization on X-ray \cite{Wang2017ChestXray8, Tang2018attention} and CT images \cite{Shin2016inter, Hofmanninger2015cvpr}. This approach boosts the size of datasets and label sets. However, current studies can only extract image-level labels, which cannot be accurately mapped to specific lesions on the image. The DeepLesion dataset\footnote{\url{https://nihcc.box.com/v/DeepLesion}} consists of lesions from a variety of body parts on CT images. It has been adopted to train algorithms for universal lesion detection \cite{Yan20183DCE}, retrieval \cite{Yan2018LesGraph}, segmentation and measurement \cite{Cai2018seg,Tang2018recist}. This paper will explore its usage on lesion-level semantic annotation. Another line of study directly generates reports according to the whole image \cite{Wang2018TieNet, Zhang2017MDNet}. Although the generated reports may learn to focus on certain lesions on the image, it is difficult to assess the usability of generated reports. The key information in reports are the labels. If we can accurately predict the labels for each lesion on the image, the creation of high-quality (structured) reports would be straightforward. 

{\bf Multilabel image classification:} Multilabel image classification \cite{Zhang2014Survey} is a long-standing topic that has been tackled from multiple angles. A direct idea is to treat each label independently with a binary cross-entropy loss \cite{Wang2017ChestXray8}. The pairwise ranking loss is applied in \cite{Weston2011WSABIE, Gong2013WARP, Li2017Pairwise} to make the scores of positive labels larger than those of the negative ones for each sample. CNN-RNN \cite{Wang2016CNNRNN} uses a recurrent model to predict multiple labels one-by-one. It can implicitly model label dependency and avoid the score thresholding issue. 
In \cite{Dong2018Imbalanced}, deep metric learning and hard example mining are combined to deal with imbalanced labels.

Noisy and incomplete training labels often exist in datasets mined from the web \cite{Chua2009NUS}, which is similar to our labels mined from reports. Strategies to handle them include data filtering \cite{Krause2016noisy}, noise-robust losses \cite{Reed2015noisy}, noise modeling \cite{Misra2016noisy}, finding reliable negative samples \cite{Li2003PU}, and so on. We use a text-mining module to filter noisy positive labels and leverage label relation to find reliable negative labels. Label relations have been exploited by researchers to improve image classification. Novel loss functions were proposed in \cite{Demyanov2017tree} for labels with hierarchical tree-like structures. 
In \cite{Marino2017More, Hu2016Relation}, prediction scores of different labels are propagated between network layers whose structure is designed to capture label relations. We apply label expansion and RHEM strategies to use label relations explicitly, and at the same time employ a score propagation layer to learn them implicitly.

\section{Label Mining and Ontology}
\label{sec:dataset}


\subsection{Ontology Construction}
\label{subsec:ontology}

We constructed our lesion ontology based on RadLex \cite{Langlotz2006RadLex}, a comprehensive lexicon for standardized indexing and retrieval of radiology information resources \cite{NIH2016RadLex, BioPortal2018RadLex}. The labels in our lesion ontology can be categorized into three classes: {\bf 1.\ Body parts}, which include coarse-level body parts (e.g., chest, abdomen), organs (lung, lymph node), fine-grained organ parts (right lower lobe, pretracheal lymph node), and other body regions (porta hepatis, paraspinal); {\bf 2.\ Types}, which include general terms (nodule, mass) and more specific ones (adenoma, liver mass); and {\bf 3.\ Attributes}, which describe the intensity, shape, size, etc., of the lesions (hypodense, spiculated, large). 

\begin{figure}[]
	\begin{center}
		\includegraphics[width=.8\linewidth,trim=190 200 430 130, clip]{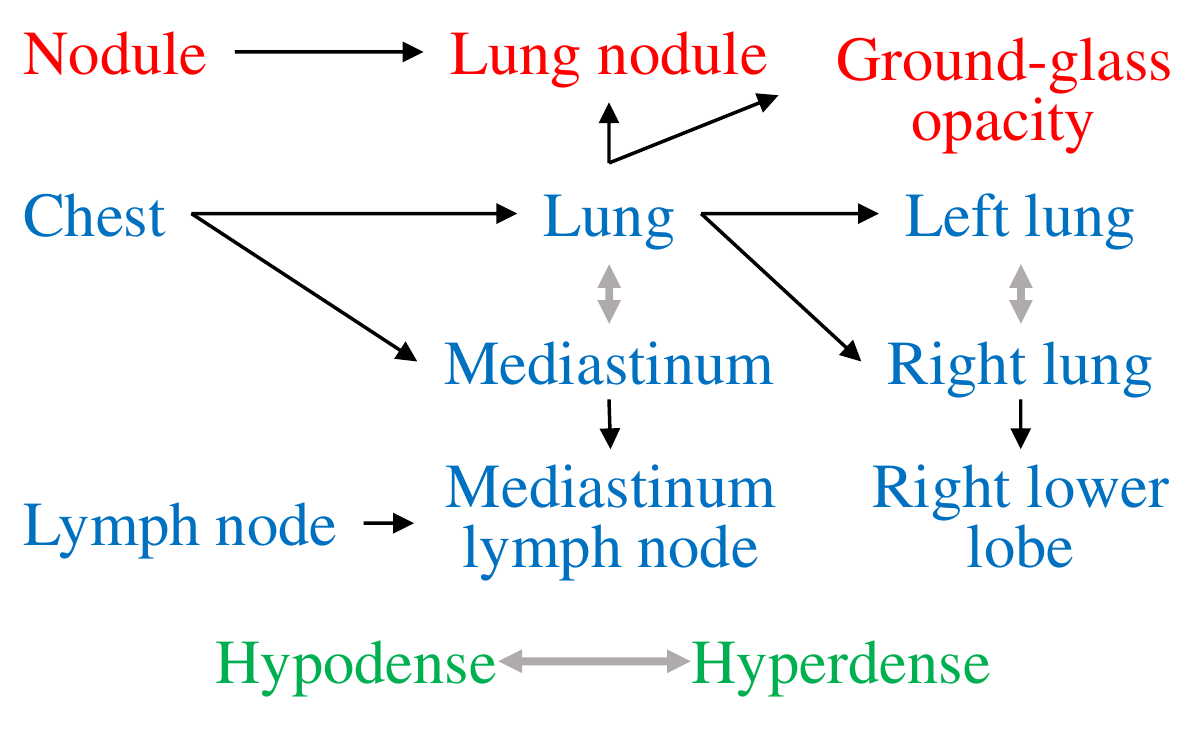} 
	\end{center}
	\caption{Sample labels with relations. Blue, red, and green labels correspond to body parts, types, and attributes, respectively. Single-headed arrows point from the parent to the child. Double-headed arrows indicate exclusive labels.}
	\label{fig:example_label_graph}
\end{figure}

The labels in the lesion ontology are organized in a hierarchical structure (\Fig{example_label_graph}). For example, a fine-grained body part (left lung) can be a part of a coarse-scale one (lung); a type (hemangioma) can be a sub-type of another one (neoplasm); and a type (lung nodule) can locate in a body part (lung). These relations form a directed graph instead of a tree, because one child (lung nodule) may have multiple parents (lung, nodule). Some labels are also mutually exclusive, meaning that the presence of one label signifies the absence of others (e.g., left and right lungs). However, in \Fig{example_label_graph}, chest and lymph node are not exclusive because they may physically overlap; lung nodule and ground-glass opacity are not exclusive either since they may coexist in one lesion. We hypothesize that if labels $ a $ and $ b $ are exclusive, any child of $ a $ and any child of $ b $ are also exclusive. This rule can help us in annotation of exclusive labels.

\subsection{Relevant Label Extraction}
\label{subsec:label_extract}

After constructing the lesion ontology, we extracted labels from the associated radiology reports of DeepLesion \cite{Yan2018DeepLesion}. In the reports, radiologists describe the lesions and sometimes insert hyperlinks, size measurements, or slice numbers (known as bookmarks) in the sentence to refer to the image of interest. In this work, we only used the sentences with bookmarks to text-mine labels associated with the lesions. First, we tokenized the sentence and lemmatized the words in the sentence using NLTK \cite{Bird2016NLTK} to obtain their base forms. Then, we matched the named entity mentions in the preprocessed sentences and normalized them to labels based on their synonyms.


The bookmarked sentences often contain a complex mixture of information describing not only the bookmarked lesion but also other related lesions and unrelated things. A sample sentence is shown in \Fig{overall_framework}, where the word ``BOOKMARK'' is the hyperlink of interest, while ``OTHER\_BMK'' is the hyperlink for another lesion. There are 4 labels matched based on the ontology, namely large, nodule, right lower lobe, and right middle lobe. Among them, ``right lower lobe'' is irrelevant since it describes another lesion. In other examples, there are also uncertain labels such as ``adenopathy or mass''. Since both the irrelevant and uncertain labels may bring noise to downstream training, we developed a text-mining module to distinguish them from relevant labels. Specifically, we reformulate it as a relation classification problem. Given a sentence with multiple labels and bookmarks, we aim to assign relevant labels to each bookmark from all label-bookmark pairs.

To achieve that, we propose to use a CNN model based on Peng et al.~\cite{peng2018extracting,Peng2019ichi}. The input of our model consists of two parts: the word sequence with mentioned labels and bookmarks, and the sentence embedding \cite{chen2018biosentvec}. The model outputs a probability vector corresponding to the type of the relation between the label and the bookmark (irrelevant, uncertain, and relevant). Due to space limit, we refer readers to \cite{Peng2019ichi} for details about this algorithm.



\section{Lesion Annotation Network (LesaNet)}
\label{sec:label_annot}

\begin{figure*}[]
	\begin{center}
		\includegraphics[width=\linewidth,trim=0 40 0 55, clip]{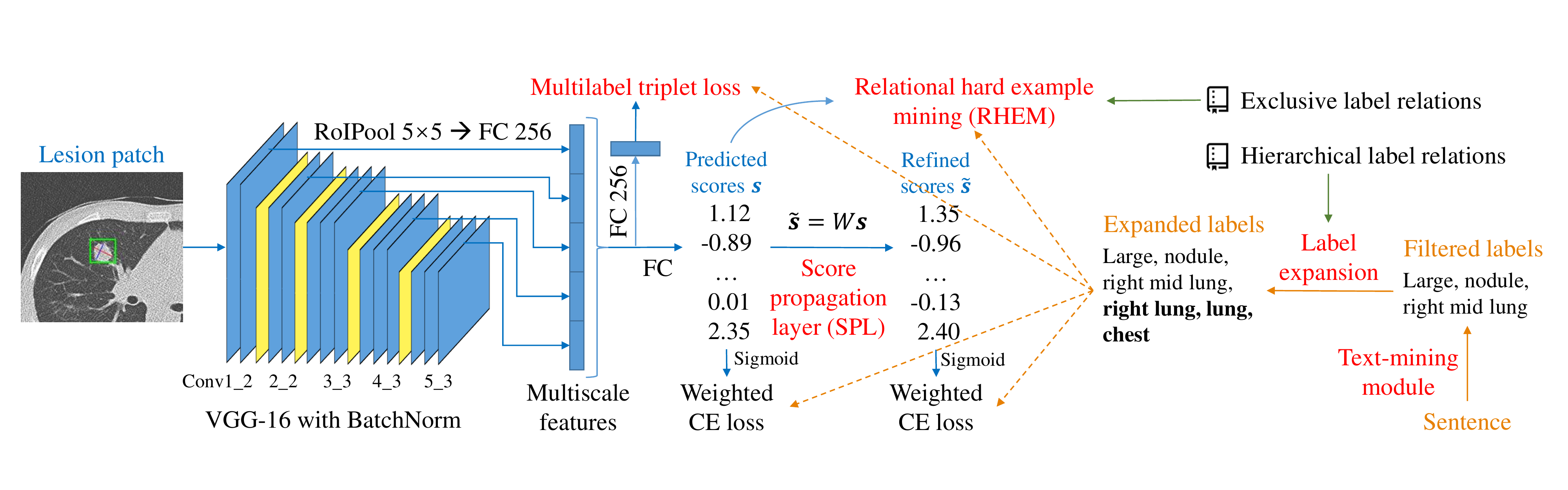} 
	\end{center}
	\caption{The framework of LesaNet. The input is the lesion image patch and the final output are the refined scores $ \tilde{\bm{s}} $. The expanded labels are used to train LesaNet and optimize the four losses. Modules in red are our main contributions.}
	\label{fig:LesaNet_framework}
\end{figure*}

\Fig{LesaNet_framework} displays the framework of the proposed lesion annotation network (LesaNet). In this section, we introduce each component in detail.

\subsection{Multiscale Multilabel CNN}
\label{subsec:basic_cnn}

The backbone of the network is VGG-16 \cite{Simonyan2015Vgg} with batch normalization \cite{Ioffe2015BN}. In our task, different labels may be best modeled by features at different levels. For instance, body parts require high-level contextual features while many attributes depict low-level details. Therefore, we use a multiscale feature representation similar to \cite{Yan2018LesGraph}. Region of interest pooling layers (RoIPool) \cite{Girshick2015fast} are used to pool the feature maps to $5\times5$ in each convolutional block. For conv1\_2, conv2\_2, and conv3\_3, the RoI is the bounding-box of the lesion in the patch to focus on its details. For conv4\_3 and conv5\_3, the RoI is the entire patch to capture the context. Each pooled feature map is then projected to a 256D vector by a fully-connected layer (FC) and concatenated together. After another FC layer, the network outputs a score vector $ \bm{s}\in\mathbb{R}^C $, where $ C $ is the number of labels. Because positive cases are sparse for most labels, we adopt a weighted cross-entropy (CE) loss \cite{Wang2017ChestXray8} for each label as in \Eq{loss_wce}, where $ B $ is the number of lesion images in a minibatch; $ \sigma_{i,c} = \text{sigmoid}(s_{i,c})$ is the confidence of lesion $ i $ having label $ c $, whose ground-truth is $ y_{i,c} \in\{0,1\}$; the loss weights are $ \beta_c^\text{p}=|P_c+N_c|/|2P_c|, \beta_c^\text{n}=|P_c+N_c|/|2N_c| $, where $ P_c,N_c $ are the numbers of positive and negative cases of label $ c $ in the training set, respectively.
\vspace{-1mm}
\begin{equation}\label{eq:loss_wce}
L_{\text{WCE}} = \sum_{i=1}^B\sum_{c=1}^C \left(\beta_c^\text{p} y_{i,c} 
	\log\sigma_{i,c} + \beta_c^\text{n}(1-y_{i,c}) \log(1-\sigma_{i,c}) \right).
\end{equation}

\subsection{Leveraging Label Relations}
\label{subsec:label_rel}

{\bf Label expansion:} Labels extracted from reports are not complete. The hierarchical label relations can help us infer the missing parent labels. If a child label is true, all its parents should also be true. In this way, we can find the labels ``right lung'', ``lung'', and ``chest'' in \Fig{LesaNet_framework} based on the existing label ``right mid lung'' in both training and inference.

{\bf Relational hard example mining (RHEM):} Label expansion cannot complete other missing labels if their children labels are not mentioned in the report. This problem occurs when radiologists did not describe every attribute of a lesion or omitted the fine-grained body part. Although it is hard to retrieve these missing positive labels, we can utilize the exclusive relations to find reliable negative labels. In other words, if the expanded labels of a lesion are reliably 1, then their exclusive labels should be reliably 0.

One challenge of our task is that some labels are difficult to learn. We hope the loss function to emphasize them automatically. Inspired by online hard example mining (OHEM) \cite{Shrivastava2016OHEM}, we define the online difficulty of label $ c $ of lesion $ i $ as:
\vspace{-1mm}
\begin{equation}\label{eq:shrem_diffi}
\delta_{i,c}=|\sigma_{i,c}-y_{i,c}|^\gamma,
\end{equation}
$ \gamma>0 $ is a focusing hyper-parameter similar to the focal loss \cite{Lin2017Focal}. Higher $ \gamma $ puts more focus on hard examples. Then, we sample $ S $ lesion-label pairs according to $ \delta $ in the minibatch, and compute their average CE loss. The higher $ \delta_{i,c} $ is, the more times it will be sampled. Hence, the loss will automatically focus on hard lesion-label pairs. This stochastic sampling strategy works better in our experiments than the selection strategy in OHEM \cite{Shrivastava2016OHEM} and the reweighting one in focal loss \cite{Lin2017Focal}. An important note is that the sampling is only performed on reliable lesion-label pairs, so as to avoid treating missing positive labels as hard negatives. RHEM also works as a dynamic weighting mechanism for imbalanced labels, thus there is no need to impose weights \cite{Shrivastava2016OHEM} as the $ \beta $ in \Eq{loss_wce}. We combine the CE loss of RHEM with \Eq{loss_wce} instead of replacing it, since some labels have no exclusive counterparts and have to be learned from \Eq{loss_wce}.

\subsection{Score Propagation Layer}
\label{subsec:score_propa}

A score propagation layer (SPL) is attached at the end of LesaNet (\Fig{LesaNet_framework}). It is a simple FC layer that refines the predicted scores with a linear transformation matrix $ W $, followed by a weighted CE loss (\Eq{loss_wce}). $ W $ is initialized with an identity matrix and can learn to capture the first-order correlation between labels. Although the hierarchical and exclusive label relations have been explicitly expressed by label expansion and RHEM, it is still useful to have SPL as it can enhance the scores of positively related labels and suppress the scores of labels with negative correlation and clear separation. On the other side, some exclusive labels can be very similar in location and appearance, for instance, hemangioma and metastasis in liver. When SPL sees a high score of hemangioma, it will know that it may also be a metastasis since in some cases they are hard to distinguish. Therefore, SPL will actually increase the score for metastasis slightly instead of suppressing it. This mechanism is particularly beneficial to improve the recall of rare labels whose prediction scores are often low. This rationale distinguishes SPL from previous knowledge propagation methods \cite{Hu2016Relation} that enforce negative weights on exclusive labels, which led to a lower performance in our task. By observing the learned $ W $, we can also discover more label correlation and compare them with our prior knowledge. 

\subsection{Multilabel Triplet Loss}
\label{subsec:triplet}

Interpretability is important for CAD tasks \cite{Sahiner2018survey}. We expect the algorithm to provide evidence for its predictions. After classifying a lesion, it is desirable if LesaNet can show lesions in the database that have similar labels, which will help the user better understand its prediction as well as the lesion itself. This is a joint lesion annotation and retrieval problem. Lesion retrieval was studied in \cite{Yan2018LesGraph}, but only 8 coarse-scale body part labels were used. In this paper, we use the comprehensive labels mined from reports to learn a feature embedding to model the similarity between lesions. As shown in \Fig{LesaNet_framework}, an FC layer is applied to project the multiscale features to a 256D vector, followed by a triplet loss \cite{schroff2015facenet}. To measure the similarity between two images with multiple labels, Zhao et al.\ \cite{Zhao2015retrieval} used the number of common positive labels as a criterion. However, we argue that each lesion may have a different number of labels, so the number of disjoint positive labels also matters. Suppose $ X$ and $Y $ are the set of positive labels of lesions $ A$ and $B $, we use the following similarity criterion:
\begin{equation}\label{eq:sim_crit}
	\text{sim}(A,B) = |X\cap Y|^2/|X\cup Y|.
\end{equation}
When training, we first randomly sample an anchor lesion $ A $ from the minibatch, and then find a similar lesion $ B $ from the minibatch so that $ \text{sim}(A,B)\geq\theta $, finally find a dissimilar lesion $ C $ so that $ \text{sim}(A,C)<\text{sim}(A,B) $. $ \theta $ is the similarity threshold. We sample $ T $ such triplets from the minibatch and calculate the triplet loss:
\begin{equation}\label{eq:triplet_loss}
	L_{\text{triplet}}=\frac{1}{T}\sum_{t=1}^T \max(0,\, d(A,B)-d(A,C)+\mu),
\end{equation}
where $ d(A,B) $ is the L2 distance of the embeddings of $ A $ and $ B $, $ \mu $ is the margin. $ L_{\text{triplet}} $ makes lesions with similar label sets closer in the embedding space.

The final loss of LesaNet combines the four components:
\begin{equation}\label{eq:final_loss}
	L = L_{\text{WCE}} + L_{\text{CE, RHEM}} + L_{\text{WCE, SPL}} + \lambda L_{\text{triplet}}.
\end{equation}

\section{Experiments}
\label{sec:exp}

\subsection{Dataset}
\label{subsec:exp_data}

From DeepLesion and its associated reports, we gathered 19,213 lesions with sentences as the training set, 1,852 as the validation set, and 1,759 as the test set. Each patient was assigned to one of the subsets only. The total number is smaller than DeepLesion because not all lesions have bookmarks in the reports. We extracted labels as per Sec.\ \ref{subsec:label_extract}, then kept the labels occurring at least 10 times in the training set and 2 times in both the validation (val) and the test sets, resulting in a list of 171 unique labels. Among them, there are 115 body parts, 27 types, and 29 attributes. We extracted hierarchical label relations from RadLex followed by manual review and obtained 137 parent-child pairs. We further invited a radiologist to annotate mutually exclusive labels and obtained 4,461 exclusive pairs.

We manually annotated the label relevance (relevant / uncertain / irrelevant, Sec.\ \ref{subsec:label_extract}) in the val and test sets with two expert radiologists' verification. As a result, there are 4,648 relevant, 443 uncertain, and 1,167 irrelevant labels in the test set. The text-mining module was trained on the val set and applied on the training set. Then, labels predicted as relevant or uncertain in the training set were used to train LesaNet. LesaNet was evaluated on the relevant labels in the test set. Because the bookmarked sentences may not include all information about a lesion, there may be missing annotations in the test set when relying on sentences only. Hence, two radiologists further manually annotated 500 random lesions in the test set in a more comprehensive fashion. On average, there are 4.2 labels per lesion in the original test set, and 5.4 in the hand-labeled test set. An average of 1.2 labels are missing in each bookmarked sentence. We call the original test set the ``text-mined test set'' because the labels were mined from reports. The second hand-labeled test set is also used to evaluate LesaNet.

\subsection{Implementation Details}
\label{subsec:implement}


For each lesion, we cropped a $ 120 \text{mm}^2 $ patch around it as the input of LesaNet. To encode 3D information, we used 3 neighboring slices to compose a 3-channel image. Other details about the dataset and image preprocessing are presented in Sec.\ \ref{subsec:data}. For the weighted CE loss, we clamped the weights $ \beta $ to be at most 300 to ensure training stability. For RHEM, we set $ \gamma=2 $ and $ S=10^4 $. For the triplet loss, we empirically set $ \theta=1, \mu=0.1$, and $ T=5000$. The triplet loss weight was $ \lambda=5 $ since this loss is generally smaller than other loss terms. LesaNet was implemented using PyTorch \cite{pytorch} and trained from scratch. Lesions with at least one positive label were used in training. The batch size was 128. LesaNet was trained using stochastic gradient descent (SGD) with a learning rate of 0.01 for 10 epochs, then with 0.001 for 5 more epochs.

\subsection{Evaluation Metric}
\label{subsec:eval_metric}

The AUC, i.e.\ the area under the receiver operating characteristic (ROC) curve, is a popular metric in CAD tasks \cite{Wang2017ChestXray8, Cheng2016Breast}. However, AUC is a rank-based metric and does not involve label decision, thus cannot evaluate the quality of the final predicted label set in the multilabel setting. Thus, we also computed the precision, recall, and F1 score for each label, which are often used in multilabel image classification tasks \cite{Zhang2014Survey}. Each metric was averaged across labels with equal weights (per-class-averaging). Overall-averaging \cite{Zhang2014Survey} was not adopted because it biases towards the frequent labels (chest, abdomen, etc.)~which are less informative. To turn confidence scores into label decisions, we calibrated a threshold for each label that yielded the best F1 on the validation set, and then apply it on the test set.


\subsection{Lesion Annotation Results}
\label{subsec:exp_img}

\begin{table*}[]
	\centering
	\begin{tabular}{l|cccc|cccc}
		\hline
		\multirow{2}{*}{Method} & \multicolumn{4}{|c|}{Text-mined test set} & \multicolumn{4}{|c}{Hand-labeled test set} \\
		\cline{2-9}
		& AUC & Precision & Recall & F1	& AUC & Precision & Recall & F1 \\
		\hline
		Multiscale multilabel CNN	&  0.9048	&  0.2738	&  0.5224	&  0.2823	&	0.9151	&	0.3823	&	0.5340    &	0.3894	\\ 
		WARP \cite{Gong2013WARP} 	&	0.9250 &	0.2441 &	\bf 0.6202 &	0.3017 &	0.9316 &	 \bf 0.6677 &	0.3273 &	0.3325 \\ 
		Lesion embedding \cite{Yan2018LesGraph} 	& 0.8933	&	0.2290 &	0.5767 &	0.2610 &	0.9017 &	0.3496 &	\bf 0.5776 &	0.3615 \\ 
		LesaNet	&  \bf 0.9344	&  0.3593	&  0.5327	&  \bf 0.3423	&	\bf 0.9398	&	0.4737	&	0.5274	&	\bf 0.4344		\\
		\hline
		w/o score propagation layer		&  0.9275	&  \bf 0.3680	& \color{red} \underline{0.4733}	&  0.3233	&	0.9326	&	0.4833	&	\color{red} \underline{0.4965}	&	\color{red} \underline{0.4092}
		\\
		w/o RHEM	&  0.9338	& \color{red} \underline{0.2983}	&   0.5550	& \color{red} \underline{0.3178}	&	0.9374	&	\color{red} \underline{0.4341}	&	0.5327	&	0.4303
		\\
		w/o label expansion	& \color{red} \underline{0.9148} 	&  0.3523	&  0.5104	&  0.3270	&	\color{red} \underline{0.9236}	&	0.4503	&	0.5420	&	0.4205
		\\
		w/o	text-mining module	& 0.9334	&  0.3365	& 0.5350	&  0.3324	&	0.9392	&	0.4869	&	0.5361	&	0.4250
		\\
		w/o triplet loss	&  0.9312	&  0.3201	&  0.5394	&  0.3274	&	0.9335	&	0.4645	&	0.5624	&	0.4337  \\
		\hline
	\end{tabular}
	\caption{Multilabel classification accuracy averaged across labels on two test sets. Bold results are the best ones. Red underlined results in the ablation studies are the worst ones, indicating the ablated strategy is the most important for the criterion.}
	\label{tbl:acc}
\end{table*}


A comparison of different methods and an ablation study of our method are shown in \Table{acc}. The baseline method is the multiscale multilabel CNN described in Sec.\ \ref{subsec:basic_cnn}. The weighted approximate ranking pairwise loss (WARP) \cite{Gong2013WARP} is a widely-used multilabel loss that aims to rank positive labels higher than the negative ones. We applied it to the multiscale multilabel CNN. We defined that fine-grained labels should rank higher than coarse-scale ones if they are all positive. Lesion embedding \cite{Yan2018LesGraph} was trained on DeepLesion based on labels of coarse-scale body parts, lesion location, and size. Among these four methods, LesaNet achieved the best AUC and F1 scores on the two test sets.

The AUCs in \Table{acc} are relatively high. The algorithms have correctly ranked most positive cases higher than negative ones, proving the effectiveness of the algorithms. However, the F1 scores are relatively low. There are mainly two reasons: 1) The dataset is highly imbalanced with many rare labels. A total of 78 labels have fewer than 10 positive cases in the text-mined test set. These labels may have many more false positives (FPs) than true positives (TPs) when testing, resulting in a low F1. 2) There are missing annotations in the test sets, which is why the accuracies (especially precisions) on the hand-labeled set are significantly higher than the text-mined test set.

\begin{table}
	\centering
	\setlength{\tabcolsep}{3pt}
	\begin{tabular}{lrrc|clrr}
		\hline 
		Label &  AUC & F1 &&& Label & AUC & F1 \\
		\hline
		Chest & 96.2 & 90.2 &&& Nodule & 89.1 & 66.9 \\
		Lung & 98.6 & 92.0 &&& Cyst & 96.0 & 40.7 \\
		Liver & 98.6 & 78.8 &&& Adenoma & 99.9 & 30.8 \\
		Lymph node & 93.7 & 76.2 &&& Metastasis & 74.0 & 10.7 \\
		Adrenal gland & 99.5 & 76.2 &&& Hypodense & 87.7 & 50.9 \\
		Right mid lung & 98.7 & 56.6 &&& Sclerotic & 99.7 & 75.4 \\
		Pancreatic tail & 97.5 & 35.3 &&& Cavitary & 94.9 & 25.0 \\
		Paraspinal & 97.5 & 9.8 &&& Large & 80.6 & 17.5 \\
		\hline
	\end{tabular}
	\caption{Accuracies (\%) of typical body parts, types, and attributes.}
	\label{tbl:acc_per_term}
\end{table}

Accuracies of some typical labels on the text-mined test set are displayed in \Table{acc_per_term}. 
The average AUC of body parts, types, and attributes are 0.9656, 0.9044, and 0.8384, respectively. Body parts are easier to predict since they typically have more regular appearances. The visual feature of some labels (e.g., paraspinal, nodule) is variable, thus harder to learn. The high AUC and low F1 of ``paraspinal'' can be explained by the lack of positive test cases (see the explanation in the last paragraph). Some types (e.g., metastasis) can be better predicted by incorporating additional prior knowledge and reasoning. Attributes have lower AUCs partially because some attributes are subjective (``large'') or can be subtle (``sclerotic''). Besides, radiologists typically do not describe every attribute of a lesion in the report, thus there are missing annotations in the test set.

\begin{figure*}[]
	\begin{center}
		\includegraphics[width=.95\linewidth,trim=80 250 370 100, clip]{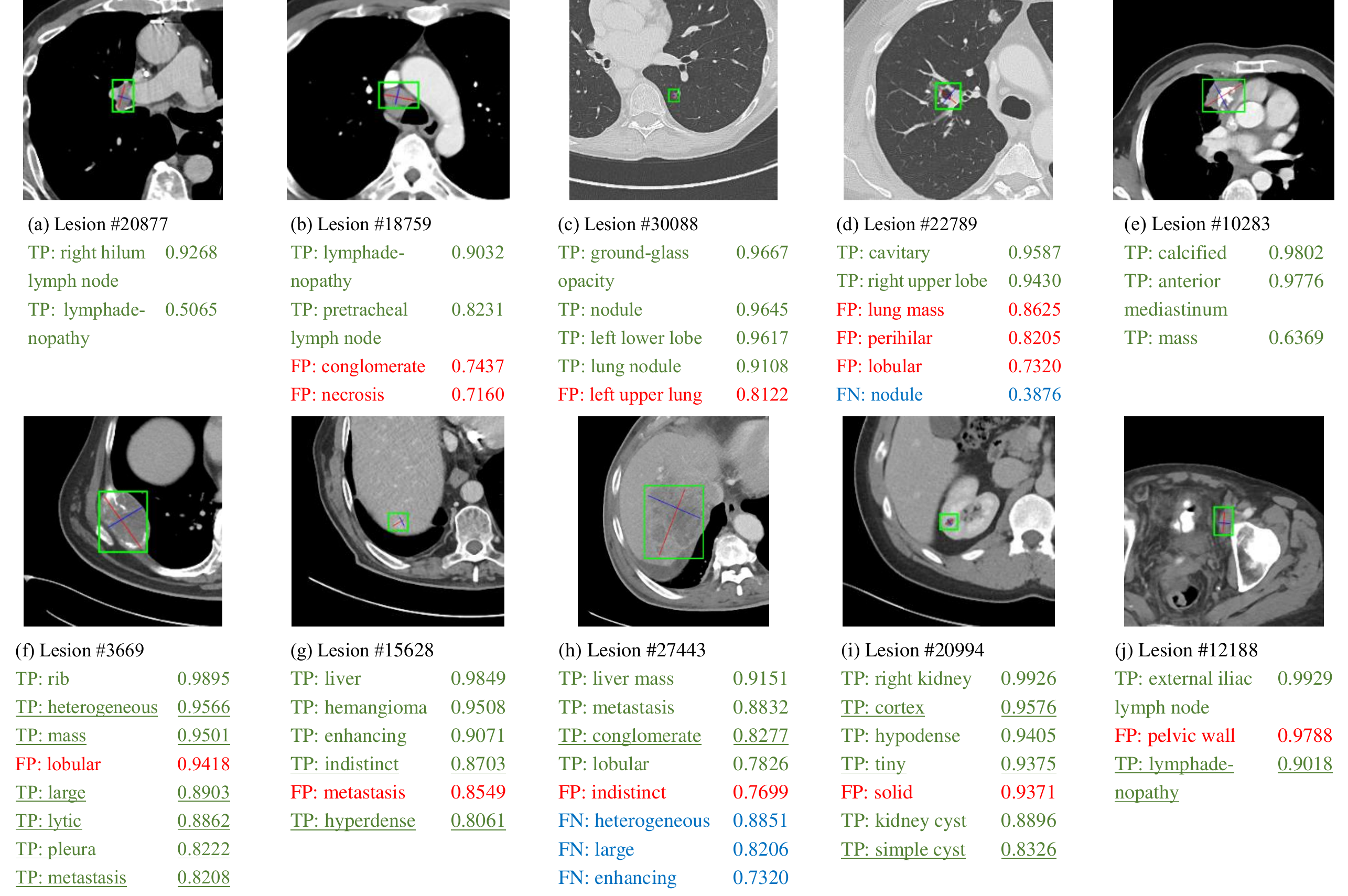} 
	\end{center}
	\caption{Sample predicted labels with confidence scores on the text-mined test set. Green, red, and blue results correspond to TPs, FPs, and FNs (false negatives), respectively. Underlined labels are TPs with missing annotations, thus were treated as FPs during evaluation. Only the most fine-grained predictions are shown with their parents omitted for clarity. }
	\label{fig:show_lesion_landscape}
\end{figure*}

\Fig{show_lesion_landscape} demonstrates examples of our predictions. LesaNet accurately predicted the labels of many lesions. For example, in subplots (a) and (b), two fine-grained body parts (right hilum and pretracheal lymph nodes) were identified; In (c) and (d), a ground-glass opacity and a cavitary lung lesion; In (g) and (h), a hemangioma and a metastasis in liver. Some attributes were also predicted correctly, such as ``calcified'' in (e), ``lobular'' in (h), and ``tiny'' in (i). Errors can occur on some similar body parts and types. In (c), although ``left lower lobe'' has a high score, ``left upper lung'' was also predicted, since the two body parts are close. In (g), ``metastasis'' is a wrong prediction as it may be hard to be distinguished from hemangioma in certain cases. Some rare and / or variable labels were not learned very well, such as ``conglomerate'' and ``necrosis'' in (b). Please see the supplementary material for more results.

\begin{figure}[]
	\begin{center}
		\includegraphics[width=.95\linewidth,trim=0 0 0 0, clip]{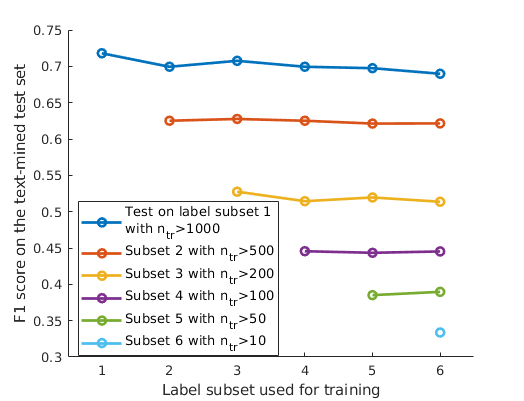} 
	\end{center}
	\caption{Accuracy of training and testing LesaNet on different subsets of labels. Each curve corresponds to a test subset.}
	\label{fig:acc_subset_label}
	\vspace{-3mm}
\end{figure}

It is efficient to jointly learn all labels holistically. Furthermore, our experiments showed that it does not affect the accuracy of single labels. We conducted an experiment to train and test LesaNet on subsets of labels. For example, subset 1 consists of labels with more than 1000 occurrences in the training set ($ n_{\text{tr}}>1000 $); Subset 2 contains labels with $ n_{\text{tr}}>500 $. When trained on subset 2, we can test on both subsets 1 and 2 to see if the accuracy on subset 1 has degraded. The results are exhibited in \Fig{acc_subset_label}. We can see that for the same test set, the F1 score did not change significantly as the number of training labels increased. Thus, with more data harvested, we may safely add more clinically meaningful labels into training. On the other hand, as more rare labels were added to the test set, the F1 became lower. Fine-grained body parts, types, and many attributes are rare. They are harder to learn due to the lack of training cases. Possible solutions include harvesting more data automatically \cite{Yan2018DeepLesion} and using few-shot learning \cite{Sung2018Fewshot}.

\subsection{Ablation Study and Analysis}
\label{subsec:components}

\begin{figure}[]
	\begin{center}
		\includegraphics[width=\linewidth,trim=0 0 0 0, clip]{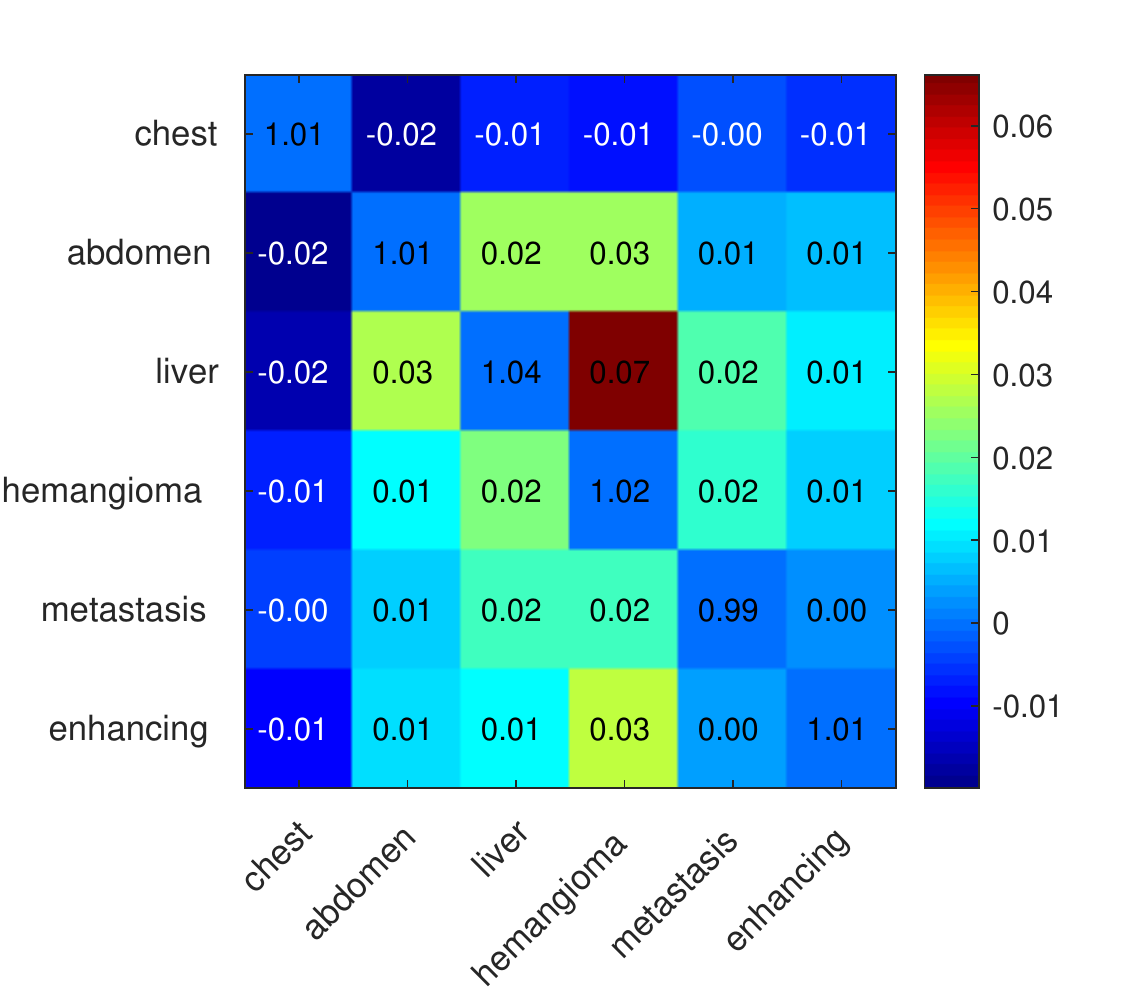} 
	\end{center}
	\vspace{-2mm}
	\caption{A part of the learned score propagation weights $ W $. The weight in row $ i $, column $ j $ is $ w_{ij} $, i.e., the refinement of label $ i $'s score received from label $ j $'s score. The final scores $ \tilde{\bm{s}}=W\bm{s} $.}
	\vspace{-3mm}
	\label{fig:score_propa_wts}
\end{figure}

{\bf Score propagation layer:} From the ablation study in \Table{acc}, we find that removing SPL decreased the average per-class recall by 3\%. Among it, the recall of frequent labels ($ n_{\text{tr}}>1000 $) only decreased 0.4\%, showing that SPL is important for the recall of rare labels, at the cost of small precision loss. We further examined the learned transformation matrix $ W $ in SPL, see \Fig{score_propa_wts} for an example. We can find that $ W(\text{liver, hemangioma}) $ and $ W(\text{enhancing, hemangioma}) $ are high. It means SPL discovered the fact that if a lesion is a hemangioma in DeepLesion, it is highly likely in the liver and enhancing, so SPL increased the scores for ``liver'' and ``enhancing''. In turn, the scores of liver and enhancing also contributed positively to the final score of hemangioma (see \Fig{show_lesion_landscape} (g) for an example of hemangioma). Note that these relations were not explicitly defined in the ontology. The label ``chest'' is exclusive with ``abdomen'' and ``liver'', so the learned weights between them are negative. As explained in Sec.\ \ref{subsec:score_propa}, hemangioma and metastasis in the liver are hard for the algorithm to distinguish, so SPL also learned positive weights between them. In the future, using our holistic and comprehensive prediction framework, we may try to incorporate more human knowledge into the model, such as ``type $ a $ locates in body part $ b $ and has attribute $ c $'', ``type $ d $ is similar to type $ e $ except for attribute $ f $''.

{\bf Relational hard example mining:} RHEM, on the contrary to SPL, is crucial for improving the precision (\Table{acc}), probably because it suppressed the scores of the reliable hard negative labels at the cost of mildly decreased recall. In RHEM, the hard negative labels of a lesion were selected from the exclusive labels of existing positive ground-truths. If we discard this ``reliable'' requirement and select negative labels from all labels that are not positive, the precision will increase 1.5\% because we suppressed more negative labels, but the recall will decrease 4\% since many suppressed negative labels are actually positive due to missing annotations.

{\bf Label expansion:} Without it, the training set will lose 40\% (parent) labels, thus the accuracy was not good.

{\bf Text-mining module: } When this was not used, the overall accuracy dropped as the irrelevant training labels brought noises. However, the performance did not degrade substantially, showing that our model is able to tolerate noisy labels to a certain degree \cite{Krause2016noisy}. We also found training with the relevant + uncertain labels was better than using relevant labels only, which is because most uncertain labels are radiologists' inferences that are very likely to be true, especially if we only consider the lesion's appearance.

\begin{figure}[]
	\begin{center}
		\includegraphics[width=\linewidth,trim=0 180 420 0, clip]{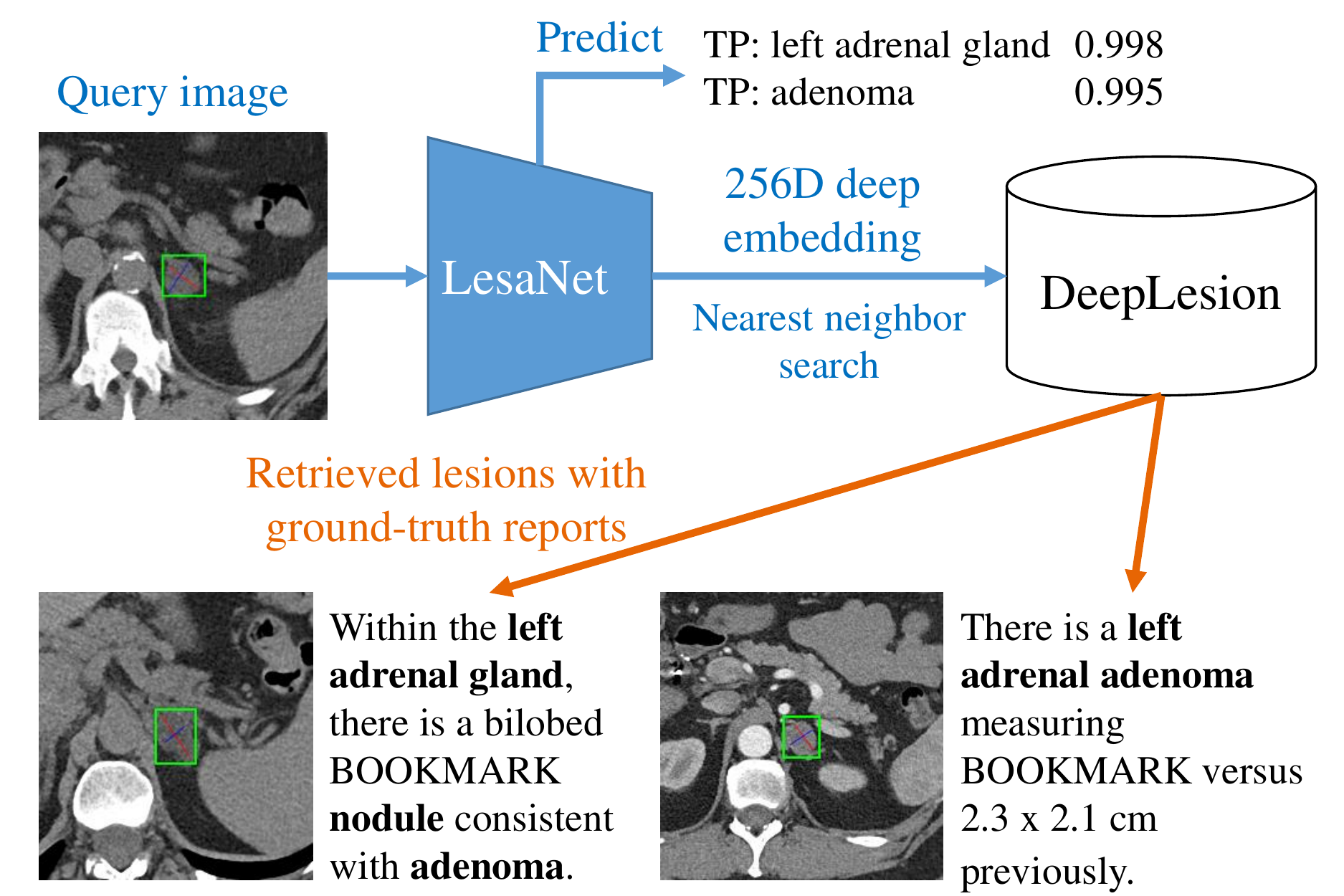} 
	\end{center}
	\caption{Sample lesion retrieval results.}
	\vspace{-2mm}
	\label{fig:retrieval_example}
\end{figure}

{\bf Triplet loss:} The triplet loss also contributed to the classification accuracy slightly. 
The 256D embedding learned from the triplet loss can be used to retrieve similar lesions from the database given a query one. In \Fig{retrieval_example}, LesaNet not only predicted the labels of the query lesion correctly, but also retrieved lesions with the same labels, although their appearances are not identical. The retrieved lesions and reports can provide evidences to the predicted labels as well as help the user understand the query lesion.

More qualitative and quantitative results are presented in the supplementary material.

\section{Conclusion and Future Work}
\label{sec:conclusion}

In this paper, we studied the holistic lesion annotation problem, and proposed a framework to automatically learn clinically meaningful labels from radiology reports and label ontology. A lesion annotation network was proposed with effective strategies that can both improve the accuracy and bring insights and interpretations. Our future work may include harvesting more data to better learn rare and hard labels and trying to incorporate more human knowledge.

\section*{Acknowledgment}
This research was supported by the Intramural Research Program of the NIH Clinical Center and National Library of Medicine. We thank NVIDIA for the donation of GPUs and Dr.\ Le Lu and Dr.\ Jing Xiao for their valuable comments.

\section{Supplementary Material}

\subsection{Dataset and Image Preprocessing Details}
\label{subsec:data}
\begin{figure}[]
	\begin{center}
		\includegraphics[width=\linewidth,trim=0 0 0 0, clip]{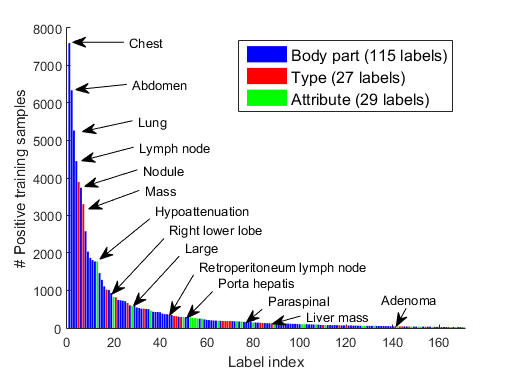} 
	\end{center}
	\caption{Distribution of label occurrences in the training set.}
	\label{fig:label_freq}
\end{figure}

The DeepLesion dataset \cite{Yan2018DeepLesion} was mined from a hospital's picture archiving and communication system (PACS) based on bookmarks, which are markers annotated by radiologists during their routine work to measure significant image findings. It is a large-scale dataset with 32,735 lesions on 32,120 axial slices from 10,594 CT studies of 4,427 unique patients. There are 1 -- 3 lesions in each axial slice. 
The numbers of training samples of some typical labels in our experiment are shown in \Fig{label_freq}. We can find that the labels are imbalanced and positive cases are sparse for most labels.

We rescaled the 12-bit CT intensity range to floating-point numbers in [0,255] using a single windowing (-1024--3071 HU) that covers the intensity ranges of the lung, soft tissue, and bone. Every image slice was resized so that each pixel corresponds to 1mm. The slice intervals of most CT scans in the dataset are either 1mm or 5mm. We interpolated in the $ z $-axis to make the intervals of all volumes 2mm.

\subsection{More Lesion Annotation Results}

\subsubsection{Examples}

\begin{figure*}
	\centering
	\includegraphics[width=.95\textwidth,trim=80 240 80 80]{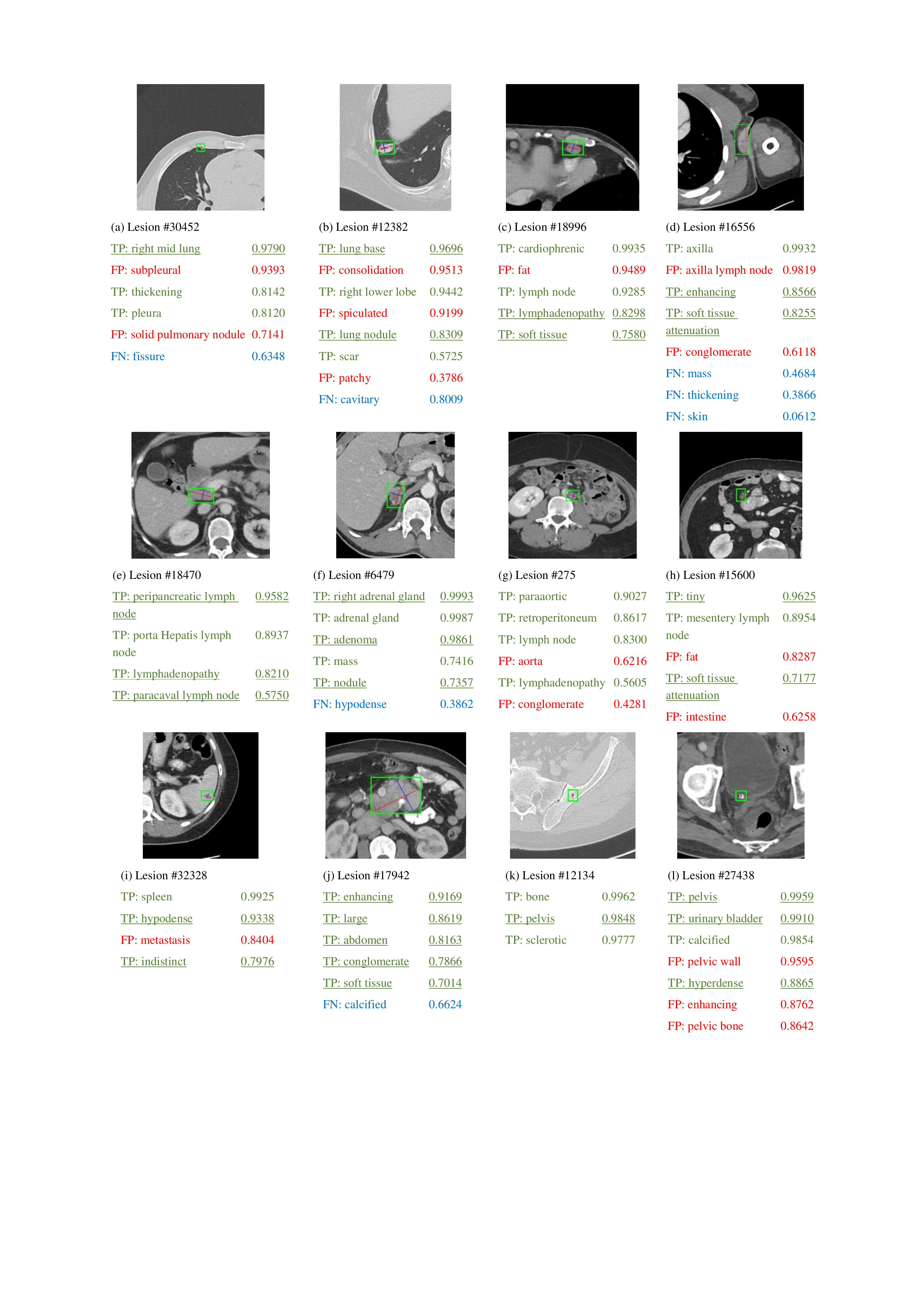} 
	\caption{Sample predicted labels with confidence scores on the text-mined test set. Green, red, and blue results correspond to TPs, FPs, and FNs, respectively. Underlined labels are TPs with missing annotations, thus were treated as FPs during evaluation. Only the most fine-grained predictions are shown with their parents omitted for clarity.}
	\label{fig:show_lesion_sup}
\end{figure*}

\Fig{show_lesion_sup} shows more lesion annotation examples of LesaNet in various body parts. We found that:

\begin{itemize}
	\item LesaNet is good at identifying fine-grained lymph nodes (subplots (c),(e),(g),(h)), which account for a major part of the DeepLesion dataset.
	\item In (d), LesaNet correctly recognized the coarse-scale body part (axilla), but it classified the lesion as a lymph node instead of a mass-like skin thickening (ground-truth). This is possibly because most axillary lesions in DeepLesion are lymph nodes, while axillary skin lesions are rare.
\end{itemize}

\subsubsection{Quantitative Results}

In order to observe the effect of the components in LesaNet more clearly, we randomly re-split the training and validation set in the patient level 10 times and rerun the ablation study. Mean and standard deviation accuracies are reported in \Table{acc_std}. Similar conclusions can be drawn from the table compared to Sec.\ 5.5 of the main paper.

The batch size during training may affect results because of the triplet loss and RHEM strategies used in LesaNet. We tested various batch sizes from 16 to 200 with
or without the two strategies. No significant correlation was observed between the settings of batch size and
accuracy. Methods with triplet loss and RHEM were consistently better than those without them.

\begin{table*}[]
	\centering
	\setlength{\tabcolsep}{3pt}
	\begin{tabular}{l|cccc|cccc}
		\hline
		\multirow{2}{*}{Method} & \multicolumn{4}{|c|}{Text-mined test set} & \multicolumn{4}{|c}{Hand-labeled test set} \\
		\cline{2-9}
		& AUC & Precision & Recall & F1	& AUC & Precision & Recall & F1 \\
		\hline
		LesaNet	&  $93.24_{0.08}$	& $30.89_{1.23}$	& $53.74_{1.62}$	& $31.76_{0.90}$	& $\bf 93.83_{0.18}$	& $47.01_{2.09}$	& $54.63_{1.41}$	& $\bf 42.29_{1.08}$		\\
		w/o score propagation layer		&  $92.42_{0.09}$	& $\bf 34.25_{2.60}$	& $\color{red} \underline{49.61_{1.55}}$	& $30.89_{0.83}$	& $\color{red} \underline{93.28_{0.30}}$	& $\bf 50.60_{2.06}$	& $\color{red} \underline{51.74_{1.72}}$	& $41.09_{1.09}$ 	\\
		w/o RHEM	&  $93.21_{0.10}$	& $\color{red} \underline{28.40_{1.49}}$	& $\bf 56.05_{2.19}$	& $31.02_{0.93}$	& $93.62_{0.22}$	& $\color{red} \underline{43.09_{1.49}}$	& $\bf 57.65_{2.11}$	& $42.04_{1.06}$		\\
		w/o label expansion	& $\color{red} \underline{92.37_{0.12}}$	& $30.16_{1.72}$	& $55.68_{1.95}$	& $\color{red} \underline{30.73_{0.60}}$	& $93.32_{0.30}$	& $45.61_{2.09}$	& $55.87_{3.14}$	& $\color{red} \underline{40.94_{1.24}}$		\\
		w/o	text-mining module	& $\bf 93.27_{0.09}$	& $30.79_{1.43}$	& $53.77_{1.90}$	& $\bf 31.94_{1.16}$	& $93.68_{0.23}$	& $46.16_{2.05}$	& $54.05_{2.68}$	& $41.49_{0.65}$		\\
		w/o triplet loss	&  $93.03_{0.07}$	& $30.65_{1.94}$	& $53.91_{1.86}$	& $31.60_{1.19}$	& $93.56_{0.18}$	& $46.29_{1.30}$	& $54.73_{1.53}$	& $41.84_{1.22}$		\\
		\hline
	\end{tabular}
	\caption{Multilabel classification accuracy averaged across labels on two test sets. Bold results are the best ones. Red underlined results in the ablation studies are the worst ones, indicating the ablated strategy is the most important for the criterion. We report mean and standard deviation of accuracies calculated on 10 random data splits formatted as $\text{mean}_{~\text{std.}}.$}
	\label{tbl:acc_std}
\end{table*}

\subsection{More Lesion Retrieval Examples}

\begin{figure*}
	\centering
	\includegraphics[width=.84\textwidth,trim=70 70 70 90]{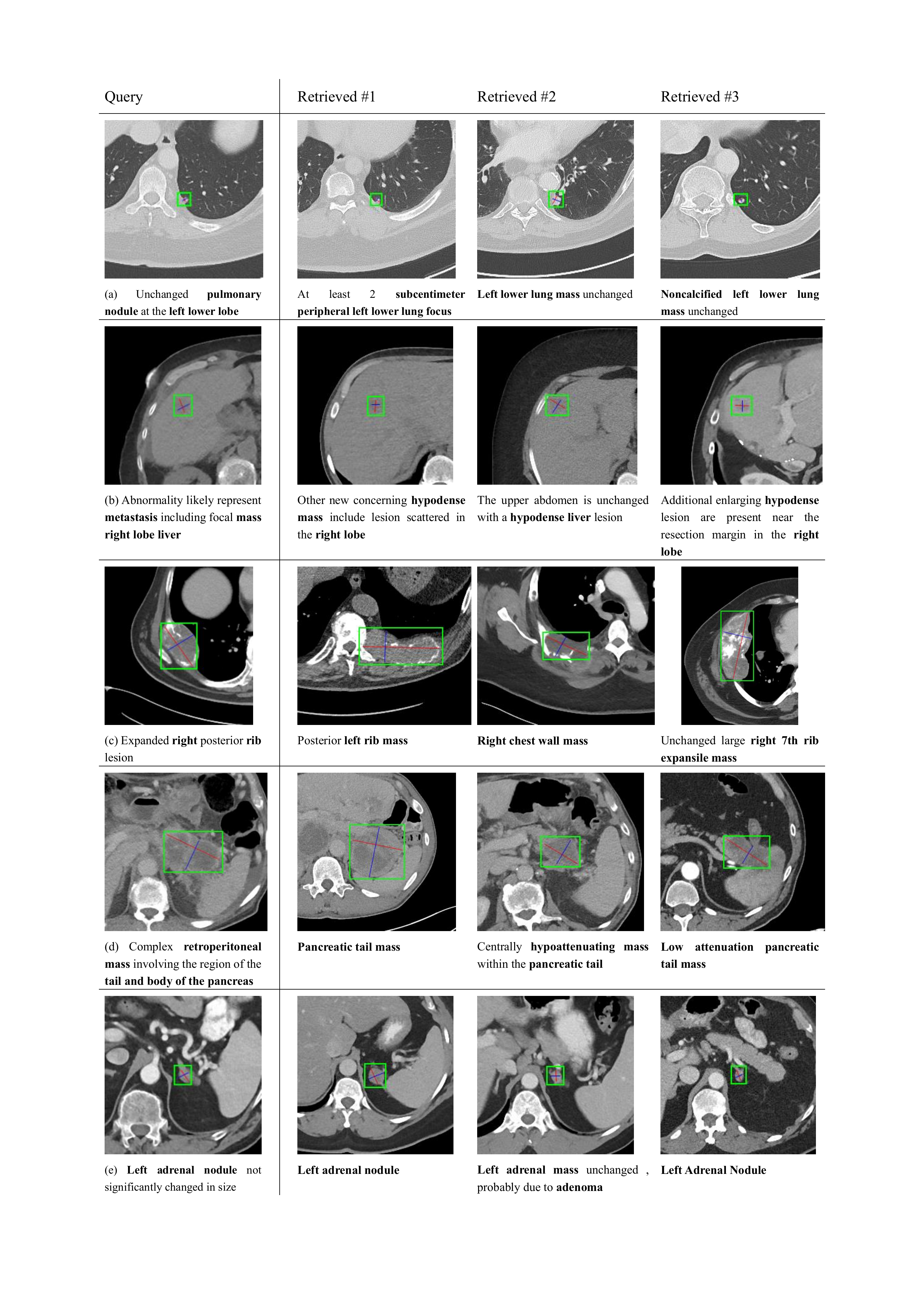} 
	\caption{Sample lesion retrieval results of LesaNet. The input of LesaNet is the lesion image patch only, whereas the associated report sentence is shown for reference. The irrelevant words in the sentences describing other lesions have been removed for clarity.}
	\label{fig:retrieval_example_sup}
\end{figure*}

\Fig{retrieval_example_sup} demonstrates more lesion retrieval examples of LesaNet (please refer to Fig.\ 7 in the main paper). We constrain that the query and all retrieved lesions must come from different patients, so as to better exhibit the retrieval ability and avoid finding identical lesions of the same patient. For lesions that are common in DeepLesion, such as lung nodules and liver masses, it is easy for LesaNet to retrieve lesions that are very similar in both visual appearance and semantic labels, \eg~\Fig{retrieval_example_sup} (a) and (b). Moreover, LesaNet is also able to retrieve lesions that look different but share similar semantic labels, \eg the rib/chest wall mass in subplot (c), the pancreatic tail mass in (d), and the left adrenal nodule in (e).

We have conducted another experiment to quantitatively compare the lesion retrieval accuracy of LesaNet and lesion embedding \cite{Yan2018LesGraph}. We used the lesions in the text-mined test set as queries to retrieve similar lesions from the training set, which has no patient-level overlap with the test set. The accuracy criterion is the average cumulative gain (ACG), which is defined as the average number of overlapping labels between the query and each of the top-K retrieved samples \cite{Zhao2015retrieval}. The ACG@top-5 of lesion embedding \cite{Yan2018LesGraph} is 2.25, meaning that a retrieved lesion shares an average of 2.25 common labels with the query lesion. The ACG@top-5 of LesaNet is 2.36.  LesaNet learned from more fine-grained labels text-mined from radiology reports, which is the main reason of its improved accuracy, despite the fact that it uses a shorter embedding vector (256D vs.\ 1024D) and was not primarily trained for retrieval.

{\small
\bibliographystyle{ieee_fullname}
\bibliography{lesion_annotation}

\begin{thebibliography}{10}\itemsep=-1pt

\bibitem{BioPortal2018RadLex}
BioPortal.
\newblock {Radiology Lexicon}, 2018.

\bibitem{Bird2016NLTK}
Steven Bird, Steven Bird, and Edward Loper.
\newblock {NLTK: The natural language toolkit}.
\newblock In {\em Annual Meeting of the Association for Computational
  Linguistics}, pages 63--70, 2016.

\bibitem{Cai2018seg}
Jinzheng Cai, Youbao Tang, Le Lu, Adam~P. Harrison, Ke Yan, Jing Xiao, Lin
  Yang, and Ronald~M. Summers.
\newblock {Accurate Weakly-Supervised Deep Lesion Segmentation using
  Large-Scale Clinical Annotations: Slice-Propagated 3D Mask Generation from 2D
  RECIST}.
\newblock In {\em MICCAI}, pages 396--404, 2018.

\bibitem{chen2018biosentvec}
Qingyu Chen, Yifan Peng, and Zhiyong Lu.
\newblock {BioSentVec}: creating sentence embeddings for biomedical texts.
\newblock {\em arXiv preprint arXiv:1810.09302}, 2018.

\bibitem{Chen2017lung}
Sihong Chen, Jing Qin, Xing Ji, Baiying Lei, Tianfu Wang, Dong Ni, and Jie~Zhi
  Cheng.
\newblock {Automatic Scoring of Multiple Semantic Attributes with Multi-Task
  Feature Leverage: A Study on Pulmonary Nodules in CT Images}.
\newblock {\em IEEE Transactions on Medical Imaging}, 36(3):802--814, Mar.
  2017.

\bibitem{Cheng2016Breast}
Jie-Zhi Cheng, Dong Ni, Yi-Hong Chou, Jing Qin, Chui-Mei Tiu, Yeun-Chung Chang,
  Chiun-Sheng Huang, Dinggang Shen, and Chung-Ming Chen.
\newblock {Computer-Aided Diagnosis with Deep Learning Architecture:
  Applications to Breast Lesions in US Images and Pulmonary Nodules in CT
  Scans}.
\newblock {\em Sci. Rep.}, 6(1):24454, 2016.

\bibitem{Chua2009NUS}
Tat-Seng Chua, Jinhui Tang, Richang Hong, Haojie Li, Zhiping Luo, and Yantao
  Zheng.
\newblock {NUS-WIDE: A Real-World Web Image Database from National University
  of Singapore}.
\newblock In {\em Proceeding of the ACM International Conference on Image and
  Video Retrieval - CIVR '09}, page~48, 2009.

\bibitem{Clark2013TCIA}
Kenneth Clark, Bruce Vendt, Kirk Smith, John Freymann, Justin Kirby, Paul
  Koppel, Stephen Moore, Stanley Phillips, David Maffitt, Michael Pringle,
  Lawrence Tarbox, and Fred Prior.
\newblock {The cancer imaging archive (TCIA): Maintaining and operating a
  public information repository}.
\newblock {\em Journal of Digital Imaging}, 26(6):1045--1057, 2013.

\bibitem{Demyanov2017tree}
Sergey Demyanov, Rajib Chakravorty, Zongyuan Ge, SeyedBehzad Bozorgtabar,
  Michelle Pablo, Adrian Bowling, and Rahil Garnavi.
\newblock {Tree-loss function for training neural networks on weakly-labelled
  datasets}.
\newblock In {\em ISBI}, pages 287--291. IEEE, apr 2017.

\bibitem{Diamant2016liver}
Idit Diamant, Assaf Hoogi, Christopher~F. Beaulieu, Mustafa Safdari, Eyal
  Klang, Michal Amitai, Hayit Greenspan, and Daniel~L. Rubin.
\newblock {Improved Patch-Based Automated Liver Lesion Classification by
  Separate Analysis of the Interior and Boundary Regions}.
\newblock {\em IEEE J. Biomed. Heal. Informatics}, 20(6):1585--1594, 2016.

\bibitem{Dong2018Imbalanced}
Qi Dong, Shaogang Gong, and Xiatian Zhu.
\newblock {Imbalanced Deep Learning by Minority Class Incremental
  Rectification}.
\newblock {\em IEEE Transactions on Pattern Analysis and Machine Intelligence},
  pages 1--14, 2018.

\bibitem{Ferreira2017nodule}
Jos{\'{e}}~Raniery Ferreira, Marcelo~Costa Oliveira, and Paulo~Mazzoncini de
  Azevedo-Marques.
\newblock {Characterization of Pulmonary Nodules Based on Features of Margin
  Sharpness and Texture}.
\newblock {\em Journal of Digital Imaging}, pages 1--13, 2017.

\bibitem{Girshick2015fast}
Ross Girshick.
\newblock {Fast r-cnn}.
\newblock In {\em Proceedings of the IEEE international conference on computer
  vision}, pages 1440--1448, 2015.

\bibitem{Gong2013WARP}
Yunchao Gong, Yangqing Jia, Thomas Leung, Alexander Toshev, and Sergey Ioffe.
\newblock {Deep Convolutional Ranking for Multilabel Image Annotation}.
\newblock {\em arXiv preprint arXiv:1312.4894}, 2013.

\bibitem{Hofmanninger2015cvpr}
Johannes Hofmanninger and Georg Langs.
\newblock {Mapping visual features to semantic profiles for retrieval in
  medical imaging}.
\newblock In {\em CVPR}, volume 07-12-June, pages 457--465, 2015.

\bibitem{Hu2016Relation}
Hexiang Hu, Guang-Tong Zhou, Zhiwei Deng, Zicheng Liao, and Greg Mori.
\newblock {Learning Structured Inference Neural Networks with Label Relations}.
\newblock In {\em CVPR}, pages 2960--2968, 2016.

\bibitem{Ioffe2015BN}
Sergey Ioffe and Christian Szegedy.
\newblock {Batch Normalization: Accelerating Deep Network Training by Reducing
  Internal Covariate Shift}.
\newblock In {\em ICML}, pages 448--456, 2015.

\bibitem{Krause2016noisy}
Jonathan Krause, Benjamin Sapp, Andrew Howard, Howard Zhou, Alexander Toshev,
  Tom Duerig, James Philbin, and Li Fei-Fei.
\newblock {The unreasonable effectiveness of noisy data for fine-grained
  recognition}.
\newblock In {\em ECCV}, pages 301--320, 2016.

\bibitem{Langlotz2006RadLex}
Curtis~P. Langlotz.
\newblock {RadLex: a new method for indexing online educational materials}.
\newblock {\em Radiographics}, 26(6):1595--1597, Nov 2006.

\bibitem{Li2003PU}
Xiaoli Li and Bing Liu.
\newblock {Learning to classify texts using positive and unlabeled data}.
\newblock In {\em IJCAI}, pages 587--592, 2003.

\bibitem{Li2017Pairwise}
Yuncheng Li, Yale Song, and Jiebo Luo.
\newblock {Improving pairwise ranking for multi-label image classification}.
\newblock In {\em CVPR}, volume 2017-Janua, pages 1837--1845, 2017.

\bibitem{Lin2017Focal}
Tsung-Yi Lin, Priya Goyal, Ross Girshick, Kaiming He, and Piotr Doll{\'{a}}r.
\newblock {Focal Loss for Dense Object Detection}.
\newblock In {\em ICCV}, pages 2980--2988, 2017.

\bibitem{Litjens2017survey}
Geert Litjens, Thijs Kooi, Babak~Ehteshami Bejnordi, Arnaud Arindra~Adiyoso
  Setio, Francesco Ciompi, Mohsen Ghafoorian, Jeroen~A.W.M. van~der Laak, Bram
  van Ginneken, and Clara~I. S{\'{a}}nchez.
\newblock {A survey on deep learning in medical image analysis}.
\newblock {\em Medical Image Analysis}, 42:60--88, dec 2017.

\bibitem{Marino2017More}
Kenneth Marino, Ruslan Salakhutdinov, and Abhinav Gupta.
\newblock {The More You Know: Using Knowledge Graphs for Image Classification}.
\newblock In {\em CVPR}, pages 20--28, 2017.

\bibitem{Misra2016noisy}
Ishan Misra, C.~Lawrence Zitnick, Margaret Mitchell, and Ross Girshick.
\newblock {Seeing through the Human Reporting Bias: Visual Classifiers from
  Noisy Human-Centric Labels}.
\newblock In {\em CVPR}, pages 2930--2939, 2016.

\bibitem{NIH2016RadLex}
The National~Institutes of Health.
\newblock {RadLex}, 2016.

\bibitem{pytorch}
Adam Paszke, Sam Gross, Soumith Chintala, Gregory Chanan, Edward Yang, Zachary
  DeVito, Zeming Lin, Alban Desmaison, Luca Antiga, and Adam Lerer.
\newblock Automatic differentiation in pytorch.
\newblock In {\em NIPS-W}, 2017.

\bibitem{peng2018extracting}
Yifan Peng, Anthony Rios, Ramakanth Kavuluru, and Zhiyong Lu.
\newblock Extracting chemical-protein relations with ensembles of svm and deep
  learning models.
\newblock {\em Database : the journal of biological databases and curation},
  2018, Jan. 2018.

\bibitem{Peng2019ichi}
Yifan Peng, Ke Yan, Veit Sandfort, Ronald~M. Summers, and Zhiyong Lu.
\newblock {A self-attention based deep learning method for lesion attribute
  detection from CT reports}.
\newblock In {\em IEEE International Conference on Healthcare Informatics},
  2019.

\bibitem{Ravishankar2016breast}
Hariharan Ravishankar, Prasad Sudhakar, Rahul Venkataramani, Sheshadri
  Thiruvenkadam, Pavan Annangi, Narayanan Babu, and Vivek Vaidya.
\newblock {Medical Image Description Using Multi-task-loss CNN}.
\newblock In {\em LABELS 2016, DLMIA 2016}, volume~1, pages 121--129, 2016.

\bibitem{Reed2015noisy}
Scott Reed, Honglak Lee, Dragomir Anguelov, Christian Szegedy, Dumitru Erhan,
  and Andrew Rabinovich.
\newblock {Training Deep Neural Networks on Noisy Labels with Bootstrapping}.
\newblock In {\em ICLR Workshop}, 2015.

\bibitem{Sahiner2018survey}
Berkman Sahiner, Aria Pezeshk, Lubomir~M. Hadjiiski, Xiaosong Wang, Karen
  Drukker, Kenny~H. Cha, Ronald~M. Summers, and Maryellen~L. Giger.
\newblock {Deep learning in medical imaging and radiation therapy}.
\newblock {\em Med. Phys.}, oct 2018.

\bibitem{schroff2015facenet}
Florian Schroff, Dmitry Kalenichenko, and James Philbin.
\newblock Facenet: A unified embedding for face recognition and clustering.
\newblock In {\em CVPR}, pages 815--823, 2015.

\bibitem{Setio2017LUNA}
Arnaud Arindra Adiyoso et~al. Setio.
\newblock {Validation, comparison, and combination of algorithms for automatic
  detection of pulmonary nodules in computed tomography images: The LUNA16
  challenge}.
\newblock {\em Medical Image Analysis}, 42:1--13, 2017.

\bibitem{Shin2016inter}
Hoo~Chang Shin, Le Lu, Lauren Kim, Ari Seff, Jianhua Yao, and Ronald Summers.
\newblock {Interleaved text/image deep mining on a large-scale radiology image
  database for Automated Image Interpretation}.
\newblock {\em Journal of Machine Learning Research},
  17(9783319429984):305--321, 2016.

\bibitem{Shin2016tmi}
Hoo-Chang Shin, Holger~R. Roth, Mingchen Gao, Le Lu, Ziyue Xu, Isabella Nogues,
  Jianhua Yao, Daniel Mollura, and Ronald~M. Summers.
\newblock {Deep Convolutional Neural Networks for Computer-Aided Detection: CNN
  Architectures, Dataset Characteristics and Transfer Learning}.
\newblock {\em IEEE Transactions on Medical Imaging}, 35(5):1285--1298, may
  2016.

\bibitem{Shrivastava2016OHEM}
Abhinav Shrivastava, Abhinav Gupta, and Ross Girshick.
\newblock {Training Region-based Object Detectors with Online Hard Example
  Mining}.
\newblock In {\em CVPR}, pages 761--769, 2016.

\bibitem{Simonyan2015Vgg}
Karen Simonyan and Andrew Zisserman.
\newblock {Very deep convolutional networks for large-scale image recognition}.
\newblock In {\em ICLR 2015}, 2015.

\bibitem{Sung2018Fewshot}
Flood Sung, Yongxin Yang, Li Zhang, Tao Xiang, Philip H~S Torr, and Timothy~M
  Hospedales.
\newblock {Learning to Compare: Relation Network for Few-Shot Learning}.
\newblock In {\em CVPR}, pages 1199--1208, 2018.

\bibitem{Tang2018recist}
Youbao Tang, Adam~P. Harrison, Mohammadhadi Bagheri, Jing Xiao, and Ronald~M.
  Summers.
\newblock {Semi-Automatic RECIST Labeling on CT Scans with Cascaded
  Convolutional Neural Networks}.
\newblock In {\em MICCAI}, pages 405--413, jun 2018.

\bibitem{Tang2018attention}
Yuxing Tang, Xiaosong Wang, Adam~P. Harrison, Le Lu, Jing Xiao, and Ronald~M.
  Summers.
\newblock {Attention-Guided Curriculum Learning for Weakly Supervised
  Classification and Localization of Thoracic Diseases on Chest Radiographs}.
\newblock In {\em International Workshop on Machine Learning in Medical
  Imaging}, pages 249--258. Springer, Cham, sep 2018.

\bibitem{Wang2016CNNRNN}
Jiang Wang, Yi Yang, Junhua Mao, Zhiheng Huang, Chang Huang, and Wei Xu.
\newblock {CNN-RNN: A Unified Framework for Multi-label Image Classification}.
\newblock In {\em CVPR}, pages 2285--2294, 2016.

\bibitem{Wang2017ChestXray8}
Xiaosong Wang, Yifan Peng, Le Lu, Zhiyong Lu, Mohammadhadi Bagheri, and
  Ronald~M. Summers.
\newblock {ChestX-ray8: Hospital-scale Chest X-ray Database and Benchmarks on
  Weakly-Supervised Classification and Localization of Common Thorax Diseases}.
\newblock In {\em CVPR}, pages 2097--2106, may 2017.

\bibitem{Wang2018TieNet}
Xiaosong Wang, Yifan Peng, Le Lu, Zhiyong Lu, and Ronald~M Summers.
\newblock {TieNet: Text-Image Embedding Network for Common Thorax Disease
  Classification and Reporting in Chest X-rays}.
\newblock In {\em CVPR}, pages 9049--9058, 2018.

\bibitem{Weston2011WSABIE}
Jason Weston, Samy Bengio, and Nicolas Usunier.
\newblock {WSABIE: Scaling Up To Large Vocabulary Image Annotation}.
\newblock In {\em IJCAI}, pages 2764--2770, 2011.

\bibitem{Yan20183DCE}
Ke Yan, Mohammadhadi Bagheri, and Ronald~M. Summers.
\newblock {3D Context Enhanced Region-based Convolutional Neural Network for
  End-to-End Lesion Detection}.
\newblock In {\em MICCAI}, pages 511--519, 2018.

\bibitem{Yan2018DeepLesion}
Ke Yan, Xiaosong Wang, Le Lu, and Ronald~M. Summers.
\newblock {DeepLesion: automated mining of large-scale lesion annotations and
  universal lesion detection with deep learning}.
\newblock {\em Journal of Medical Imaging}, 5(3), 2018.

\bibitem{Yan2018LesGraph}
Ke Yan, Xiaosong Wang, Le Lu, Ling Zhang, Adam Harrison, Mohammadhadi Bagheri,
  and Ronald Summers.
\newblock {Deep Lesion Graphs in the Wild: Relationship Learning and
  Organization of Significant Radiology Image Findings in a Diverse Large-scale
  Lesion Database}.
\newblock In {\em CVPR}, pages 9261--9270, 2018.

\bibitem{Zhang2014Survey}
Min~Ling Zhang and Zhi~Hua Zhou.
\newblock {A review on multi-label learning algorithms}.
\newblock {\em IEEE Trans. Knowl. Data Eng.}, 26(8):1819--1837, aug 2014.

\bibitem{Zhang2017MDNet}
Zizhao Zhang, Yuanpu Xie, Fuyong Xing, Mason McGough, and Lin Yang.
\newblock {MDNet: A Semantically and Visually Interpretable Medical Image
  Diagnosis Network}.
\newblock In {\em CVPR}, pages 6428--6436, 2017.

\bibitem{Zhao2015retrieval}
Fang Zhao, Yongzhen Huang, Liang Wang, and Tieniu Tan.
\newblock {Deep semantic ranking based hashing for multi-label image
  retrieval}.
\newblock In {\em CVPR}, volume 07-12-June, pages 1556--1564, 2015.

\end{thebibliography}
}

\end{document}